\documentclass[10pt,twocolumn,letterpaper]{article}

\usepackage{iccv}
\usepackage{times}
\usepackage{epsfig}
\usepackage{graphicx}
\usepackage{amsmath}
\usepackage{amssymb, bm}
\usepackage{subfigure}
\usepackage{threeparttable}
\usepackage{multirow} 
\usepackage{color}
\usepackage{xcolor}
\usepackage{colortbl}
\usepackage{booktabs} 
\usepackage[numbers]{natbib}

\usepackage{authblk}

\usepackage[linesnumbered,boxed,ruled,vlined]{algorithm2e}

\usepackage[pagebackref=true,breaklinks=true,letterpaper=true,colorlinks,bookmarks=false]{hyperref}

\newtheorem{problem}{Problem}
\newcommand{\ood}{\text{OOD}}
\newcommand{\smallSize}{\fontsize{7pt}{\baselineskip}\selectfont}
\definecolor{blockgray}{gray}{0.85}

\iccvfinalcopy 

\def\iccvPaperID{6369} 

\ificcvfinal\fi

\begin{document}

\title{Joint Distribution across Representation Space for Out-of-Distribution Detection}

\author[]{Jingwei Xu}
\author[]{Siyuan Zhu}
\author[]{Zenan Li}
\author[]{Chang Xu}

\affil[]{State Key Lab of Novel Software Technology, Nanjing University \\
	Nanjing, China
  }

\maketitle
\ificcvfinal\thispagestyle{empty}\fi

\begin{abstract}
   Deep neural networks (DNNs) have become a key part of many modern software applications. 
   After training and validating, the DNN is deployed as an irrevocable component and applied in real-world scenarios.
   Although most DNNs are built meticulously with huge volumes of training data, data in the real world still remain unknown to the DNN model,
   which leads to the crucial requirement of runtime out-of-distribution (\ood) detection.
   However, many existing approaches 1) need \ood~data for classifier training or parameter tuning, or 2) simply combine the scores of each hidden layer as an ensemble of features for \ood~detection.
   In this paper, we present a novel outlook on in-distribution data in a generative manner, which takes their latent features generated from each hidden layer as a joint distribution across representation spaces.
   Since only the in-distribution latent features are comprehensively understood in representation space, the internal difference between in-distribution and \ood~data can be naturally revealed without the intervention of any \ood~data.
   Specifically, We construct a generative model, called Latent Sequential Gaussian Mixture (LSGM), to depict how the in-distribution latent features are generated in terms of the trace of DNN inference across representation spaces.
   We first construct the Gaussian Mixture Model (GMM) based on in-distribution latent features for each hidden layer, and then connect GMMs via the transition probabilities of the inference traces.
   Experimental evaluations on popular benchmark \ood~datasets and models validate the superiority of the proposed method over the state-of-the-art methods in \ood~detection.
   \end{abstract}
   
   
   \section{Introduction}

Deep neural networks (DNNs) receive great achievements in a variety of classification tasks, \eg, image classification, object detection, semantic segmentation, and speech recognition~\cite{afouras2018deep, he2017mask,he2016deep,long2015fully}.
Deep learning, as a branch of machine learning, is to build DNN models from the training data, and deploy the well-trained model to the domain that their distributions are assumed the same as the distribution of training data.
However, data in real-world is not usually tailored to fit the assumption of identical distribution to the training data,
which makes the quality of predictive uncertainty of the DNN models unsure~\cite{lee2017training}.
The predictive uncertainty of DNN models is closely related to the problem of model-specific out-of-distribution (\ood) that distinguishes abnormal samples far away from the distribution of training data~\cite{lee2018simple}.
The latter, \ie, detecting \ood\ data for a given DNN model, is one of the essential requirements when the model prepares for deployment~\cite{amodei2016concrete}.

Hendrycks \etal~\cite{hendrycks2017baseline} revealed this model-specific \ood\ detection problem to the society and proposed a baseline method for detection through the softmax value of the target DNN model.
In recent years, the primary insight of \ood\ detection is to design a more effective metric than softmax value to distinguish data from in-distribution to \ood. 
ODIN~\cite{liang2018enhancing} applies the temperature scaling~\cite{hinton2015distilling} to pretrained DNN model to enlarge the gap between in-distribution and \ood\ data, outperforming popular strategies~\cite{shafaei2019biased}, \eg, MC-Dropout~\cite{gal2016dropout}, DeepEnsemble~\cite{bal2017simple}, and PixelCNN++,~\cite{salimans2017pixelcnn}. 
Mahalanobis~\cite{lee2018simple} provides a distance-based approach to attain this goal. 
They determine the \ood\ input by computing the Mahalanobis distance between the hidden layer outputs. 
To improve the detection efficacy, 
one often preprocesses the input data. 
For example, ODIN borrows the idea of adversarial attack~\cite{43405}, adding the perturbation to input to separate the in-distribution and \ood\ data more clearly. 

Despite the success of the existing methods~\cite{akshay2018reducing,hendrycks2019oe, vyas2018outofdistribution}, 
they often model the \ood\ detection problem as a binary classification task and take \ood\ data as the positive class, which indicates that \ood\ data is a necessary part in the training phase.
In this sense, the proposed \ood\ detectors actually perform in a discriminative manner~\cite{jebara2012machine}. 
However, the discriminative model for \ood\ detection can easily be biased to the training \ood\ dataset, which causes that the other \ood\ datasets are still mysteries to the model. 
Contrary to the discriminative model, 
the generative model is intrinsically able to measure the predictive uncertainty and detect the outliers for the given DNN model.
The rationale is that the generative model takes \ood\ detection as a one-class classification task, which is consistent with the general knowledge in classic machine learning~\cite{ben2005outlier}.
However, outlier detection in classic machine learning is often model-independent due to poor feature extraction for the (statistical) machine learning models. 
In the era of deep learning, latent features extracted from DNN models could be effective in constructing a generative \ood\ detector.

\begin{figure}[!t]
    \centering
    \includegraphics[width=6cm]{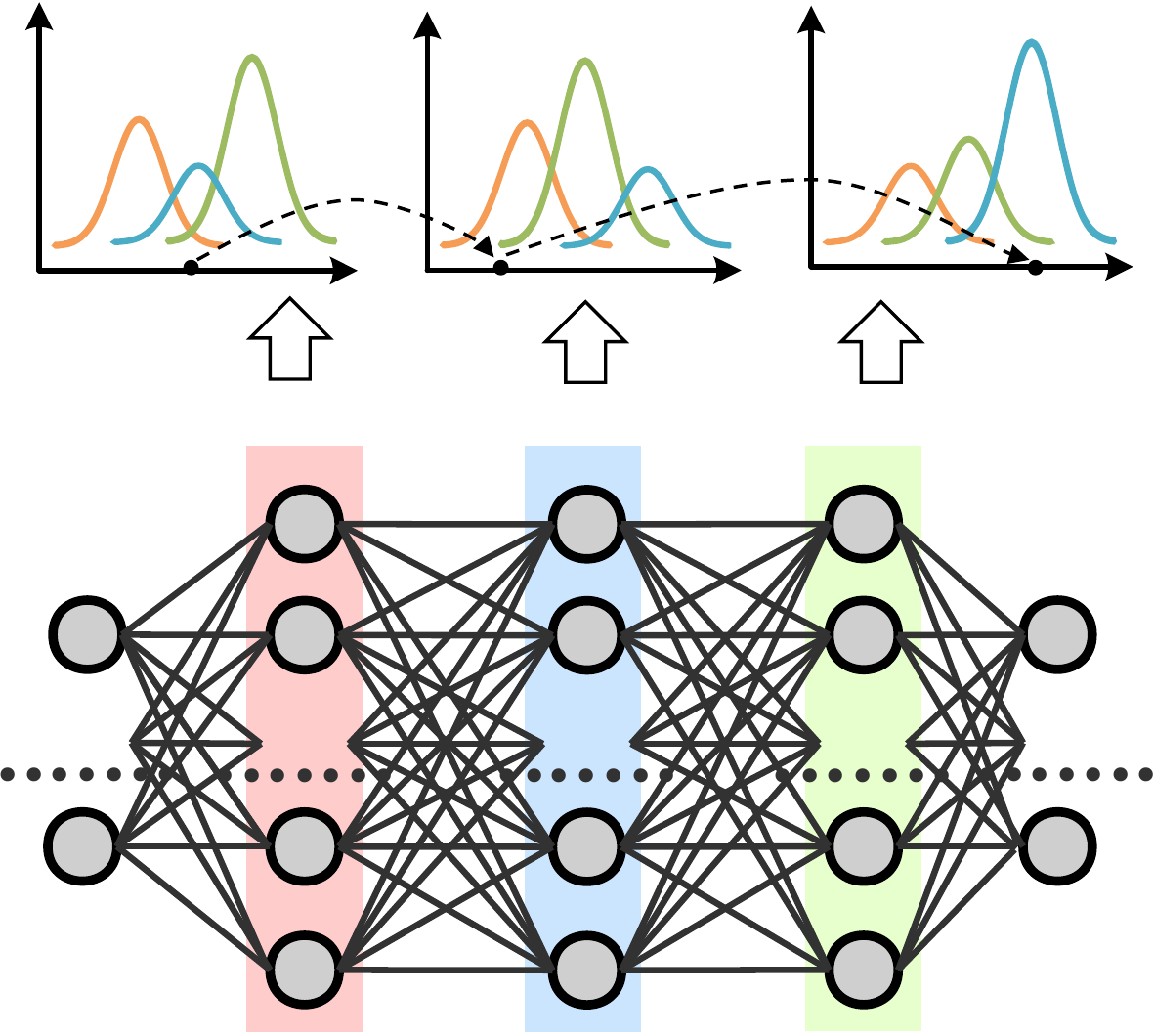}
    \caption{The illustrative example of LSGM. We construct LSGM on the highlighted three hidden layers. For each hidden layer, the distribution is built by Gaussian Mixture model, connected to the distribution of the next layer. Given an input $\bm{x}$, the trace of inference across the layers is shown in the figure.}
    \label{fig:example}
\end{figure}  

In this paper, we construct a DNN model-specific generative model, Latent Sequential Gaussian Mixture (LSGM), to connect the latent features of each representation space in series.
LSGM depicts the trace of DNN inference, \ie, recursively generates latent features in one space and transits to the location in the next space.
Based on the transitions, 
the trace of DNN inference across the representation spaces
can be revealed as a joint distribution.
Thus, 
we propose a new vision of data distribution from joint spaces rather than one single space. 
The illustrative example is shown in Fig.~\ref{fig:example}.
As a novel paradigm, LSGM is the first one to explicitly represent the process of input comprehension in the generative manner.
Specifically, an input from the in-distribution should obtain a high joint probability on the process of generating the outputs through hidden layers, whereas an \ood~input will have a low joint probability (\ie, not similar to the trace of inference for in-distribution data) in the generative model. 
In addition, as a generative model, LSGM does not need any information from \ood~data. 


We evaluate the proposed method on three kinds of noise and five real-world benchmark datasets, assuming \ood~datasets are not available for parameter tuning or learning. The experimental results show that the proposed LSGM significantly outperforms state-of-the-art methods, winning 33\footnote{Counting without regard to Generalized ODIN, since it must retrain the model, and sacrifice the prediction accuracy (drop about 3\% top-1 accuracy) for the detection.} of 35 cases in evaluation in terms of AUROC.  

The main contributions of this paper include:
\begin{itemize}
    \item A new perspective of model-specific generative model  built upon the process of DNN inference, taking 
    the inference trace into a joint distribution.
    \item A new generative probabilistic graphical model LSGM for to measure the similarity between the trace of DNN inference for the input and the in-distribution data. 
    \item An extensive analysis and evaluation in scenario of learning \ood~detector without out-of-distribution data to present the superiority of the proposed LSGM.
\end{itemize}

The rest of the paper is organized as follows. In Section 2, we present the problem statement and the background. In Section 3, we present our insight, following the description of the proposed LSGM, implementation details, and the relation to Mahalanobis. In Section 4, we present the experimental results. Finally, we conclude the paper in Section~5.

   \section{Background}


\subsection{Out-of-Distribution Detection}
The \ood\ detection problem is to identify the abnormal input for a pretrained DNN model.
Let ${P_\mathbf{X}}$ be a data distribution defined on input space $\mathcal{X}$. 
Let $\mathcal{F}$ represent a DNN model trained on a dataset drawn from the distribution ${P_\mathbf{X}}$. 
Suppose that there exists one distribution ${Q_\mathbf{X}}$ which is far away from the training distribution $P_\mathbf{X}$. 
In this case, for the DNN model $\mathcal{F}$, ${P_\mathbf{X}}$ is the in-distribution and ${Q_\mathbf{X}}$ is one of the out-of-distributions.
After the deployment of the DNN model $\mathcal{F}$, we draw a new input $\bm{x}$ from space $\mathcal{X}$. 

Based on the above notations, 
we define the following problem:
\begin{problem}{The Out-of-Distribution Detection Problem for the pretrained DNN model $\mathcal{F}$}\label{p:problem}
\begin{description}
    \item[Given:] An input $\bm{x}$ drawn from input space $\mathcal{X}$, 
    \item [Find:] The probability of that input $\bm{x}$ is in-distribution.
    \end{description}
\end{problem}

\subsection{Related Work}
As discussed in the survey~\cite{bulusu2021anomalous}, \ood\ detection falls into the category of unintentional anomaly detection.
Given a pretrained DNN model, Hendrycks~\etal analyzed that the softmax value could statistically distinguish in-distribution and \ood\ data~\cite{hendrycks2017baseline}.
Based on this observation, they used softmax value as a baseline for \ood\ detection.
ODIN~\cite{liang2018enhancing} improved the baseline with two strategies, \eg, temperature scaling and input preprocessing, to further distinguish in-distribution and \ood\ data.

Besides softmax, output features of the input in representation spaces could also be leveraged to improve the \ood\ detector. 
Some work designed distance-based metrics to distinguish \ood\ data in representation space. For example,
Mahalanobis~\cite{lee2018simple} first takes hidden layers of the DNN model as representation spaces, and then
computes Mahalanobis distance to measure how the data belongs to in-distribution in these spaces.
Some work involves additional samples to enhance the power of \ood\ detection during the DNN model training phase. 
Lee~\etal aims to train a classifier with a GAN~\cite{radford2016unsupervised, 2014Generative} to make the prediction on GAN samples with lower confidence~\cite{lee2018training}.
Outlier Exposure (OE)~\cite{hendrycks2019oe} needs a disjoint OE dataset to mimic the \ood\ dataset for testing, however the concept of disjoint between OE and \ood\ is unclear. 
To detect data from the specific~\ood, the above methods all treat the detection as a binary classification problem so that the samples from that \ood\ are necessities for training the DNN model or the detector.

\subsection{Gaussian Mixture Model for Latent Feature Generation}
When we treat \ood\ detection as a one-class problem, 
the proper way is to build a generative model for the detection.
Given a model $\mathcal{F}$, the in-distribution ${P_\mathbf{X}}$, and the target $y$, a generative model could approximate the probability of ${p(\bm{x})}$, whereas the discriminative model is limited to the estimation of conditional probability $p(y \mid \bm{x})$. 
As for \ood\ detection, it is hard to determine ${Q_\mathbf{X}}$ in practice, which makes it impossible for building a proper discriminative model.

Given an input $\bm{x}$, its latent features in representation spaces generated from the hidden layers during DNN inference can also be described by Gaussian Mixture Model, which is a popular generative model. 
Given a pretrained DNN model and the $i$-th hidden layer,
assume the in-distribution ${P_{\mathbf{X}_i}}$ on representation space $\mathcal{X}_i$
is formed by clusters, and each cluster is subject to a multivariate Gaussian distribution.
Thus, ${P_{\mathbf{X}_i}}$ could be modeled as a mixture of several multivariate Gaussian distributions, \ie, Gaussian Mixture Model.
Suppose the distribution ${P_{\mathbf{X}_i}}$ has $K_i$ clusters, the marginal distribution of $\bm{x}_i$ can be computed by
\begin{equation} \nonumber
p(\bm{x}_i)=\sum_{k=1}^K p(\bm{x}_i \mid z=k)p(z=k), 
\end{equation}
where $z$ is the latent variable with categorical distribution, indicating the probability of the clusters being selected.

During the progress of DNN inference, \ood\ data is often less comprehensively understood in representation spaces, giving a chance to detect \ood\ data via latent features.
Mahalanobis simply combines the distance-based scores in each space in a discriminative way, leaving the relations among the spaces behind.
In this paper, we assume a latent feature is generated from a Gaussian Mixture Model on its representation space, and also depend on the latent feature in the previous space.
Then,
in the view of latent features with series connection, we develop a probabilistic graphical model LSGM to construct the inference trace. Thus, the probability of inference trace estimated by LSGM can be effective in \ood\ detection.

   \section{Approach}

In this section, we first present the observations of how the deep neural network understands in-distribution data during DNN inference.
Then, based on these observations, we design the LSGM, a generative probabilistic graphical model, for latent features generated during the process of DNN inference. 
We also discuss some implementation details for LSGM, include parameter estimation, Dirichlet Process prior, and fast forward inference. 
Finally, we illustrate the relations to Mahalanobis, which is a special case of the proposed LSGM.

\subsection{Distribution of Latent Features in Representation Space}

\begin{figure}[t]
    \centering
    \subfigure[Latent features in block1]{
    \label{fig:latent1}\centering
      \includegraphics[width=0.22\textwidth]{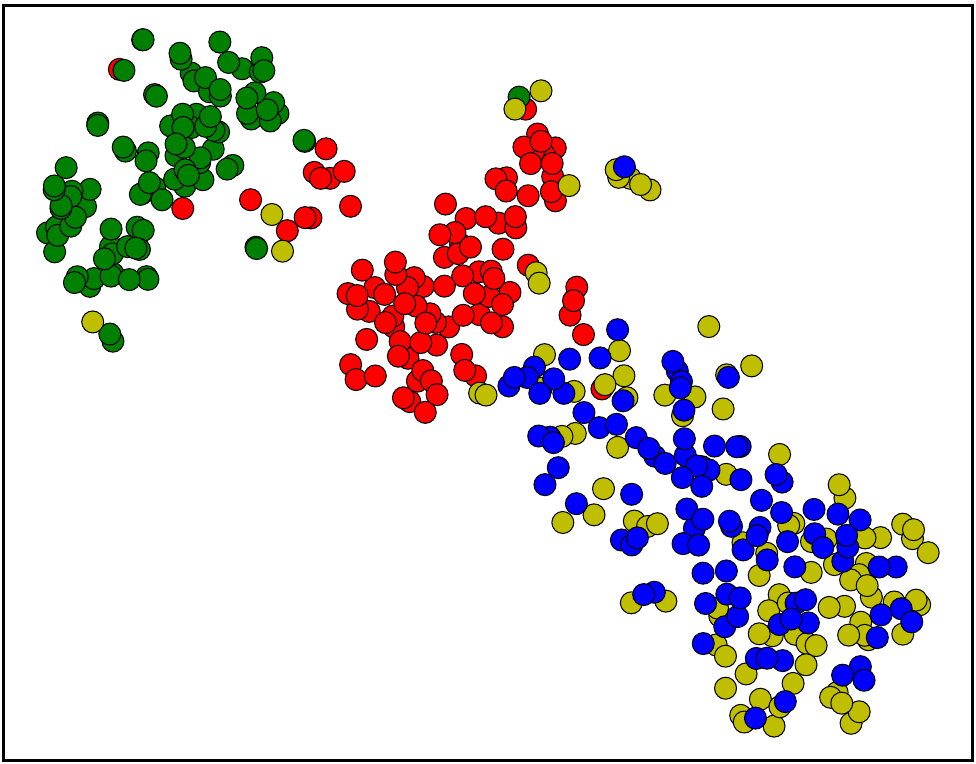}}
    \hspace{0.1in} 
    \subfigure[Latent features in block4]{
    \label{fig:penultimate}\centering
      \includegraphics[width=0.22\textwidth]{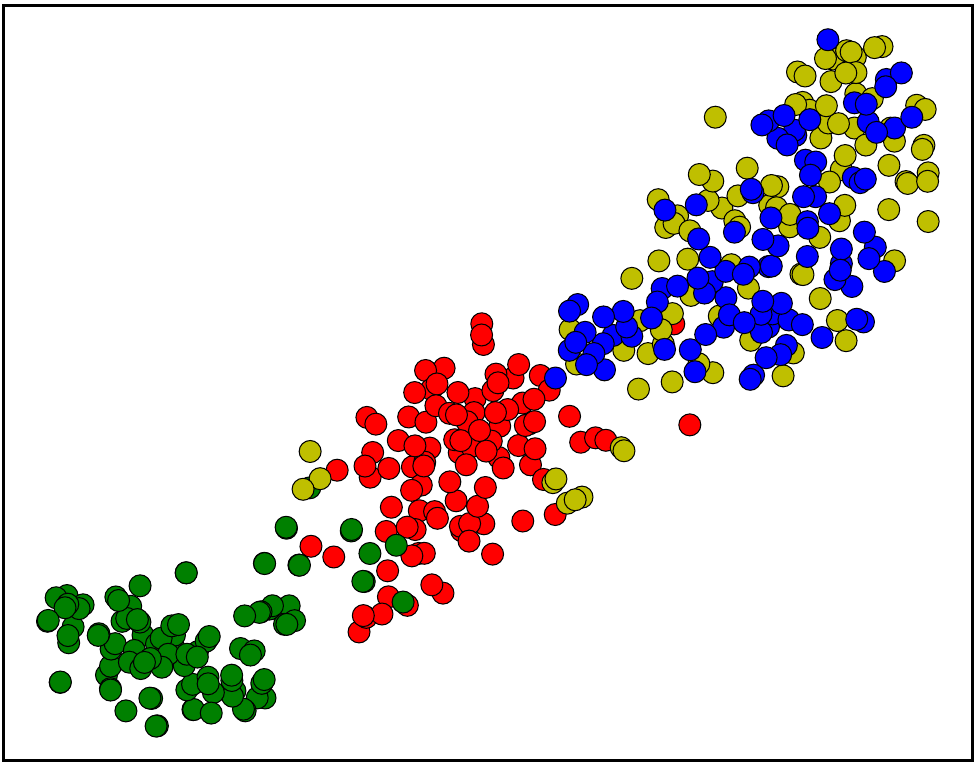}}
    \caption{Example of the first observation. Visualizations of latent features are from block1 (randomly selected) and the penultimate layer in ResNet-34 trained on CIFAR-10. In the Figure, in-distribution data (red node) are relatively closed to the center of the representation space.}
  \label{fig:latent}
  \end{figure}
  
Feature extraction is one of the intrinsic characteristics of deep neural networks.
Given a well pretrained DNN model, the corresponding training set could be comprehensively understood in representation spaces at the hidden layers of the model~\cite{goodfellow2016deep}. 
We have two observations upon the distribution of latent features to support this assumption.

The first one is \textit{the relatively high centrality of in-distribution data in representation spaces.}
Since BatchNorm is periodically assembled in many DNN models, the latent features should be surrounded in the center of the corresponding representation space.
Taking the ResNet-34 model trained on CIFAR-10 as an example, 
we randomly select 1,000 samples from in-distribution and three out-of-distribution testing datasets, respectively. 
Then, we fetch the latent features in a random hidden layer, and visualize the features via t-SNE~\cite{maaten2008visualizing}.
As shown in Fig.~\ref{fig:latent}, the in-distribution dataset is clearly separated from other \ood~datasets, with relatively centralized in the reduced two-dimensional space.

\begin{figure}[t]
    \centering
    \subfigure[In-distribution transition]{
    \label{fig:in}\centering
      \includegraphics[width=0.2\textwidth]{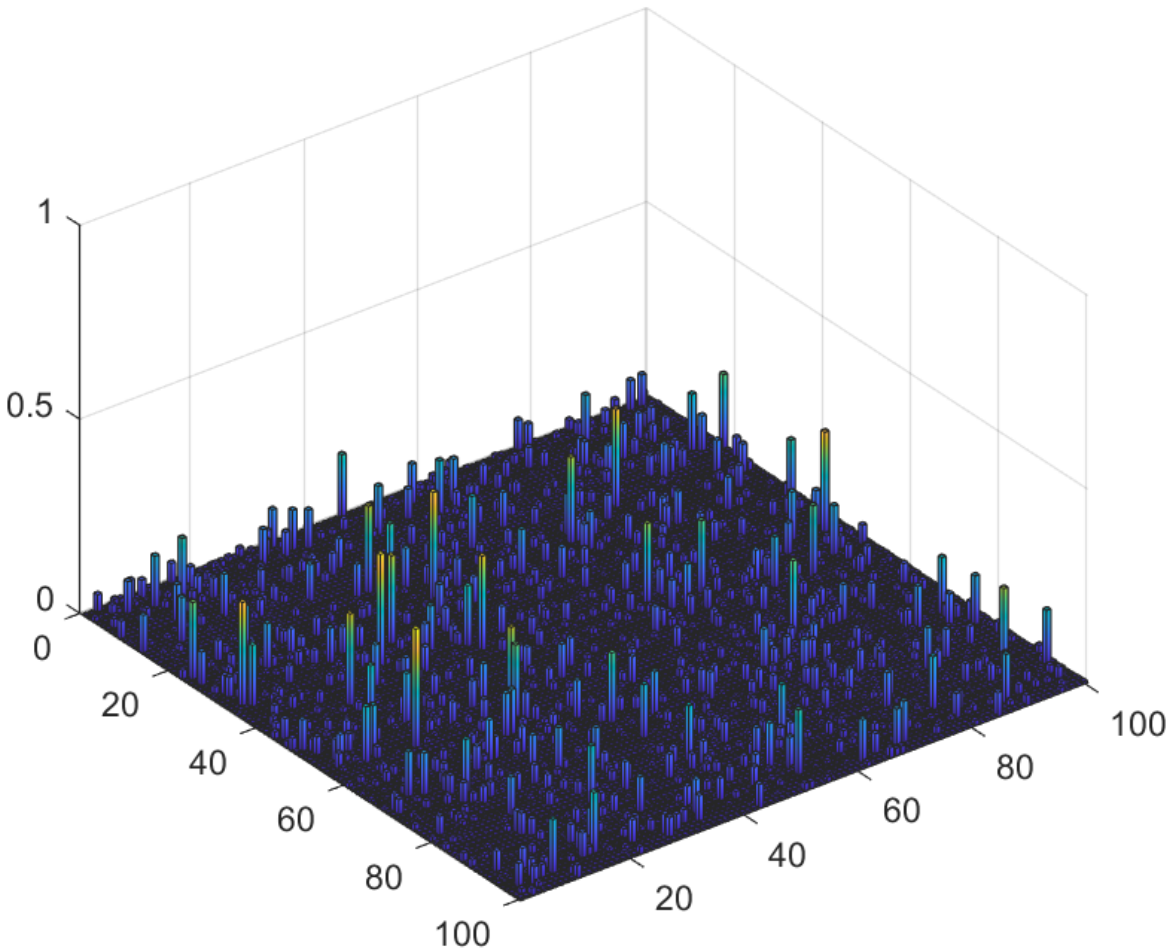}}
    \hspace{0.1in}
    \subfigure[OOD transition (SVHN)]{
    \label{fig:out}\centering
      \includegraphics[width=0.2\textwidth]{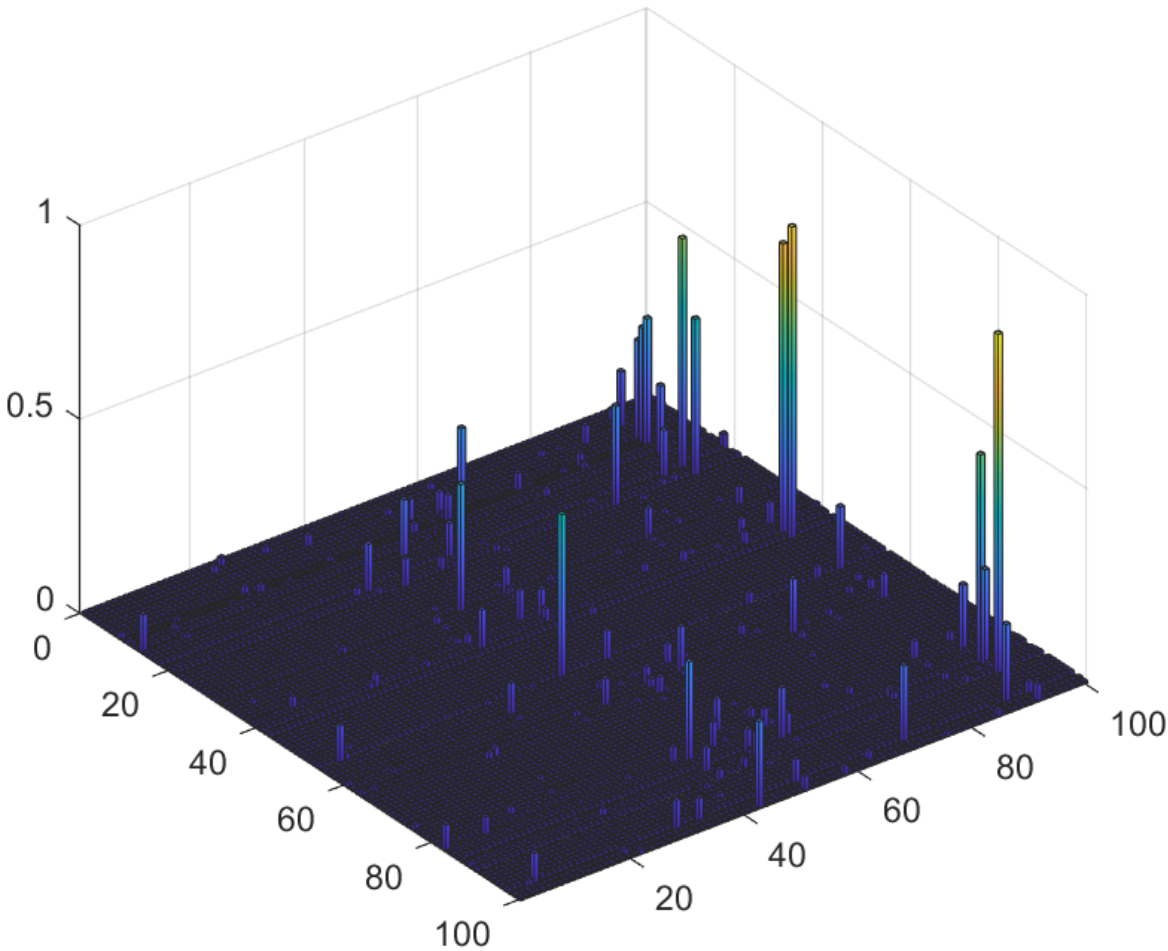}}
    \caption{Example of the second observation. Visualization for normalized statistics of transition from block2 to block3 in ResNet-34 trained on CIFAR-10.} 
  \label{fig:transition}
  \end{figure}

The second observation is that in-distribution data presents a \textit{smooth distribution of transition between two representation spaces, but the transition from the most \ood\ data is gathered in few paths}.     
Since latent features are placed in continuous space, to reveal the transitions across representation spaces, we shall first map these latent features into a discrete space.
We build Gaussian Mixture model (GMM) for latent features in the representation space to split these features into clusters. 
Then, both latent features from in-distribution and \ood\ testing data are generated by DNN inference, which is the record of the transition paths based on the flows across clusters in two adjacent spaces.
Take transitions on ResNet-34 (in-distribution: CIFAR-10 and \ood:SVHN) as an example,
the transition distributions from block2 to block3 are visualized 
in Fig.~\ref{fig:transition}. 
As shown in Fig.~\ref{fig:in}, the peaks of in-distribution transitions are numerous and low, meaning the data tends to have a smooth distribution across different paths. In contrast, \ood\ data, which is visualized in Fig.~\ref{fig:out}, concentrates on a few high peaks, demonstrating that \ood\ data tends to gather in some certain paths. The results from other layers and other DNN models share the same pattern. This phenomenon indicates that the DNN model could comprehensively understand the in-distribution data during inference, whereas they tend to present the one-sided understanding for \ood\ data.
The clear difference from transitions could further demonstrate the effectiveness of the proposed Latent Sequential Gaussian Mixture for \ood\ detection. 


Due to the two observations, we consider the centrality on representation space and discrimination of inference trace, which is a comprehensive understanding for in-distribution data v.s. one-sided understanding for \ood\ data, could naturally be expert in DNN model-specific \ood\ detection. 

\subsection{Latent Sequential Gaussian Mixture}
In this part, we present the proposed method LSGM for out-of-distribution detection.
The insight of detecting \ood\ data is to reveal the difference in the trace of DNN inference between in-distribution and \ood\ data.
Since the distributions of latent features from in-distribution data on representation spaces are constructed in DNN training phase, a novel view to the distribution of in-distribution data could be a joint distribution of these latent features on representation spaces rather than the distribution on image space. 

\begin{figure}[t]
    \centering
    \includegraphics[width=8cm]{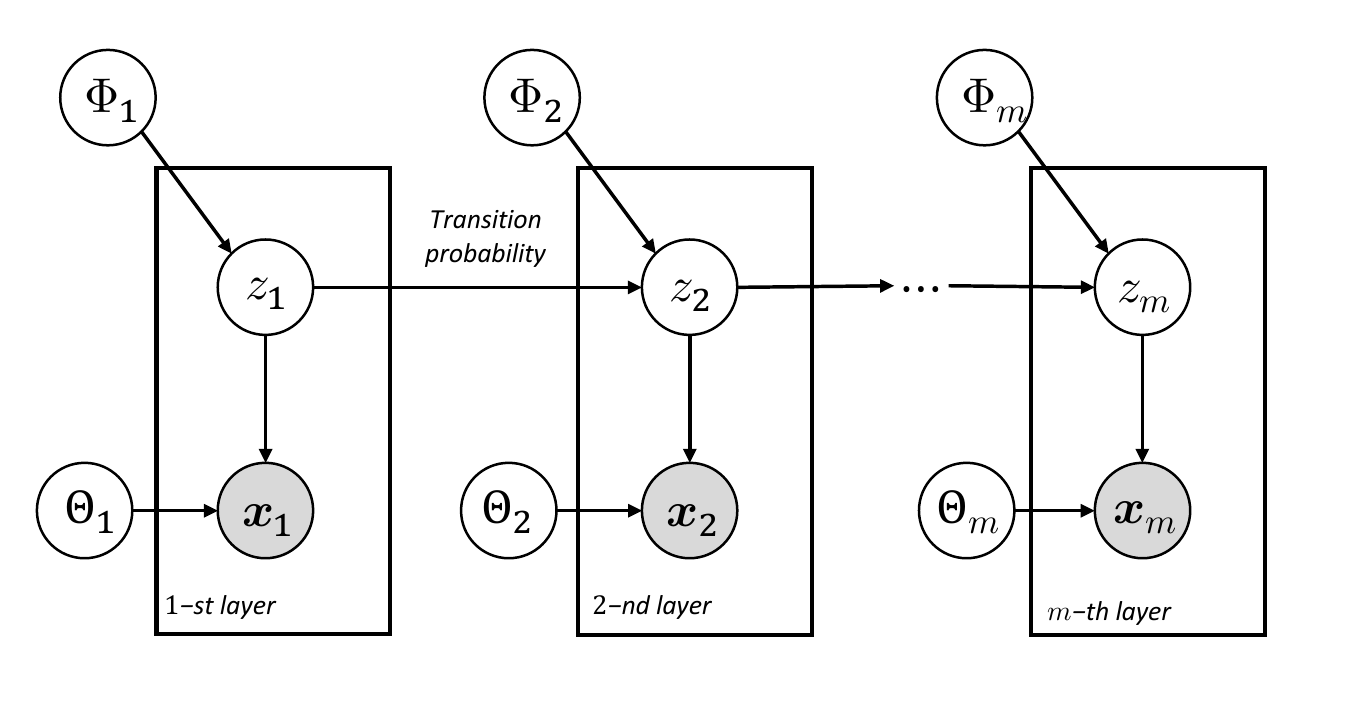}
    \caption{Probabilistic graphical model of LSGM.}
    \label{fig:lsgm}
\end{figure}

We construct a generative model that depicts how the latent features are generated in terms of the trace of DNN inference across each representation space.
The probabilistic graphical model of LSGM is shown in Fig.~\ref{fig:lsgm}, and the generative process of LSGM is as follows.

\textbf{Estimation phase:} For each $i$-th hidden layer,
\begin{itemize} 
     \item Construct GMM$_i$ with a $K_i$ dimensional Categorical distribution $\Phi_i$ and a normal distribution $\Theta_i$.
     \item Estimate parameters $\Phi_i$ and $\Theta_i$ of GMM$_i$.
     \item Estimate the transition matrix $\mathcal{P}_{i,i+1}$.
\end{itemize}

\textbf{Inference phase:} For an input $\bm{x}$, and denotes its corresponding latent features in $i$-th layer by $\bm{x}_i$, 
    \begin{itemize}
        \item Compute the mixture weights for each layer: 
        \begin{equation*}
        z_{i}\sim \text{Categorical}({\Phi}_i),
        \end{equation*}
        Let $\phi_{z_i}$ be mixture weights, \ie, the overall probability of observing a data that comes from $i$-th component.
        \item Compute the probability density of latent features $\bm{x}_i$:
            \begin{equation*} 
                \bm{x}_{i} \sim \sum_{z_i=1}^{K_i}{\phi_{z_i}}\mathcal{N}(\bm{x} \mid \bm{\mu}_{z_i},\bm{\Sigma}_{z_i}),
            \end{equation*}
	 \item Compute the joint probability of inference trace:
	 \begin{equation*}
	 P(\bm{x}_m, \dots, \bm{x}_2, \bm{x}_1 \mid \bm{\Phi}, \bm{\Theta}).
	 \end{equation*}
    \end{itemize}
    

Now we focus on how to compute the joint distribution of the latent features $\bm{x}_{1}, \bm{x}_{2},\dots, \bm{x}_{m}$ across $m$ representation spaces:
\begin{eqnarray}\label{eq:joint}
\begin{aligned}
& P(\bm{x}_{m}, \dots, \bm{x}_{2}, \bm{x}_{1} \mid \mathbf{\Phi}, \mathbf{\Theta}), \\
\end{aligned}
\end{eqnarray}
where $m$ is the number of hidden layers involved, $\mathbf{\Phi}$ and $\mathbf{\Theta}$ are the sets of $\{\Phi_1,...,\Phi_m\}$ and $\{\Theta_1,...,\Theta_m\}$, respectively.
This joint distribution statistically measures the likelihood of these latent features, which are the latent understanding of the DNN model for the given input $\bm{x}$, occurring together in its trace of DNN inference.
Apparently, according to the trace, latent features in representation spaces are not independent.
The inference process, in which the current latent features depend on the latent features on previous representation space, is like a chain process.
Consider the two connected representation spaces, to generate latent features on later {\small $(i+1)$}-th representation space, LSGM first decides which component is selected on the previous $i$-th representation space. 
Then, the occurrence of the component on later representation space is conditionally based on the previously selected component. 
After the components are selected on these two representation spaces, latent features are generated based on the component on their own representation space, respectively.
Since both $\displaystyle{z_{i}}$ and $\displaystyle{z_{i+1}}$ are categorical variables, a conditional probability table is typically used to represent the conditional distribution.

When we take an observed sequence $\bm{x}_{1}, \bm{x}_{2}, \dots, \bm{x}_{m}$ from DNN inference, all random variables $z_{1}, z_2, \dots,  z_{m}$ as hidden variables are not specified.
Thus, the joint distribution of the sequence is a marginal distribution from the joint distribution of the sequence of $\left\{\bm{x}_{1}, \dots, \bm{x}_m \right\}$ and all possible traces of $\left\{z_1,\dots,z_m \right\}$.
Take two observed $\bm{x}_{i+1}$ and $\bm{x}_{i}$ as an example, we need to sum up the terms on ${z}_i$ and ${z}_{i+1}$. 
Notice that two observed $\bm{x}_{i+1}$ and $\bm{x}_{i}$ are independent when $z_i$ and ${z}_{i+1}$ are given specific values.
The chain rule of the joint distribution of $\bm{x}_{i+1}$ and $\bm{x}_{i}$ is
    \begin{equation} \nonumber
    \begin{aligned}
    P( & \bm{x}_{i+1}, \bm{x}_{i} \mid \Phi_i, \Phi_{i+1})\\
    = & \sum_{z_i,z_{i+1}} P(\bm{x}_{i+1}, z_{i+1} \mid \bm{x}_{i}, z_{i}, \Phi_i, \Phi_{i+1}) \\
    = & \sum_{k_i,k_{i+1}} P(\bm{x}_{i+1} \mid z_{i+1}=k_{i+1}) P(\bm{x}_{i} \mid z_{i}=k_{i})\\
    &\quad  P(z_{i+1}=k_{i+1} \mid z_{i}=k_{i}, \Phi_{i+1}) P(z_{i}=k_{i} \mid \Phi_{i}).
    \end{aligned}
    \end{equation}
Since $z_{i}$ and $\Phi_{i+1}$ are independent, the conditional distribution of $z_{i+1}$ given $z_{i}$ and $\Phi_{i+1}$ could be further decomposed as
    \begin{equation} \label{eq:joint_2x_decomposed}
    \begin{aligned}
     P( & \bm{x}_{i+1}, \bm{x}_{i} \mid \Phi_i, \Phi_{i+1})\\
    = & \sum_{k_i, k_{i+1}} P(\bm{x}_{i+1} \mid z_{i+1}=k_{i+1}) P(\bm{x}_{i} \mid z_{i}=k_{i})\\
    & \quad  P(z_{i+1}=k_{i+1} \mid z_{i}=k_{i})   \\
     & \quad   P(z_{i+1}=k_{i+1} \mid \Phi_{i+1}) P(z_{i}=k_{i} \mid \Phi_{i}).
    \end{aligned}
    \end{equation}
To compute the joint distribution of Eq.~\eqref{eq:joint}, the chain rule of the joint distribution could be used for decomposition as follows:
    \begin{equation}\label{eq:joint_decomposed}
    \begin{aligned}
     P( & \bm{x}_{m}, \dots,\bm{x}_{2},\bm{x}_{1} \mid \mathbf{\Phi}, \mathbf{\Theta}) \\
    =&\sum_{\bm{z}} P(\bm{x}_{m}, z_{m} \mid \bm{x}_{m-1}, z_{m-1}) \\
    & \quad \cdots P(\bm{x}_{2}, z_2 \mid \bm{x}_{1}, z_1) P(\bm{x}_{1} \mid z_1) P(z_1),
    \end{aligned}
    \end{equation}
where $\bm{z}$ is the set of all possible traces $\{z_1, \dots, z_m\}$. 
Take the decomposed result in Eq.~\eqref{eq:joint_2x_decomposed}, we could expand the above equation in the following:
\begin{small}
\begin{eqnarray}\label{eq:joint_decomposed}
\begin{aligned}
 P( & \bm{x}_{m}, \dots, \bm{x}_{2},\bm{x}_{1} \mid \mathbf{\Phi}, \mathbf{\Theta}) \\
    = & \sum_{\bm{z}} P(\bm{x}_{m}, z_{m} \mid \bm{x}_{m-1}, z_{m-1}) \cdots P(\bm{x}_{1} \mid z_1) P(z_1),\\
=&\sum_{\bm{z}}\prod_{i=1}^{m} P(\bm{x}_{i} \mid z_{i}) P(z_{i} \mid \Phi_{i})\prod_{i=1}^{m-1} P(z_{i+1} \mid z_{i}).
\end{aligned}
\end{eqnarray}
\end{small}

The joint distribution across representation spaces is the function that measures the degree that the given input $\bm{x}$ belongs to the in-distribution data, considering the observations (\eg, latent features) from the trace of DNN inference. By leveraging the multi-spatial latent features, the data distribution is presented as a sequential model across representation spaces, with a Gaussian Mixture Model in each space.
The joint distribution is formed during the phase that the DNN model is being trained on the training set, making it proper for the representation of the in-distribution data. 
\ood\ data, which is unknown and clearly beyond the cognition of the DNN model, would show great disparity on this joint distribution compared to in-distribution data, making \ood\ data detected.


\subsection{Implementation Details}
In this part, we present parameter estimation, fast forward inference, and the Dirichlet Process prior.

\textbf{Parameter estimation}:
To use LSGM, the key inferential problem is computing the posterior distribution of the hidden variables given an observed sequence of latent features $\bm{x}_{1}, \bm{x}_{2}, \dots, \bm{x}_{m}$.
This distribution is computationally intractable in general.
Alternatively, we decompose the parameter estimation in general into several disjoint parts.
Each part represents a Gaussian Mixture Model on the corresponding representation space.  
Since each latent feature $\bm{x}_{i}$ is the observation and only depends on the hidden variable $z_i$ in terms of generative model, 
the estimation of $\Phi_i$ and $\Theta_i$ could be solved by classic variational inference.
For the transition probabilities in transition matrices, we statistically count the frequencies from existing traces in in-distribution data as the approximated estimation.

\textbf{Fast forward inference}: Assume $m$ is the number of selected hidden layers and the number of components for each hidden layer is $K_i$, the time complexity of the forward inference in Eq.~\eqref{eq:joint_decomposed} is $O(K_i^{m})$, which grows exponentially with ${\displaystyle m}$.
Inspired by the forward algorithm in Hidden Markov Model, we design the following algorithm leveraging the advantage of the conditional independence of LSGM to calculate the probability recursively. 

Consider the joint probability $ P(z_i, \bm{x}_i, \bm{x}_{1,\dots,i-1})$ represents the likelihood of $z_i$ and the current observed subsequence $\bm{x}_1$, $\bm{x}_2$, $\dots$, $\bm{x}_{i-1}$. We define the recursive function $\alpha_i(\cdot)$ with the chain rule to expand the joint probability
\begin{equation} \nonumber
    \begin{aligned}
    \alpha_i(z_i)  = & ~ P(z_i, \bm{x}_i, \bm{x}_{1,\dots,i-1})\\
    = & \sum_{z_{i-1}}^{K_{i-1}} P(z_i, \bm{x}_i, z_{i-1}, \bm{x}_{1,\dots,i-1})\\
    =& \sum_{z_{i-1}}^{K_{i-1}} P(\bm{x}_i\mid z_i, z_{i-1}, \bm{x}_{1,\dots,i-1})  \\
    & \quad  P(z_i\mid z_{i-1}, \bm{x}_{1,\dots,i-1}) P(z_{i-1},\bm{x}_{i-1}, \bm{x}_{1,\dots,i-2}).
    \end{aligned}
\end{equation}
In LSGM, $\bm{x}_i$ is conditionally independent of other variables except $z_i$, and $z_i$ is conditionally independent of other variables except $z_{i-1}$.
The above equation could be simplified as follow:
\begin{equation} \nonumber
    \alpha_i(z_i)= P(\bm{x}_i\mid z_i)\sum_{z_{i-1}}^{K_{i-1}} P(z_i\mid z_{i-1})\alpha_{i-1}(z_{i-1}).
\end{equation}
Since $ P(\bm{x}_i\mid z_i)$ is given by Gaussian Mixture Model and $ P(z_i\mid z_{i-1})$ is from the transition matrices in LSGM, we could calculate $\alpha_i(z_i)$ from $\alpha_{i-1}(z_{i-1})$ without exponential computation time. 
Assume $K_1$, $K_2$, $\dots$, $K_m$ have the same value in practice,
the time complexity of probability inference with the recursion function is $O(mK_i)$. Since $m\ll K_i$ in practice, the fast forward inference shows linearly \wrt the average number of components.

\textbf{Dirichlet Process prior}: The default distribution of latent features on each representation space is approximated via Gaussian Mixture Model.
However, Gaussian Mixture Model has several hyperparameters, \eg, number of components, that need manual selection.
Consider the infinite number of components, 
an alternative way is to apply Dirichlet Process prior as a distribution of mixture weights over Gaussian distribution components~\cite{dirichlet1999}.  
Naturally, many of these components will be redundant, and their mixing weights are close to $0$, so that we could ignore them from probability mixture.
In the implementation, the Dirichlet Process inference uses a truncated distribution with a fixed maximum number of components. 
Therefore, the actual number of components used always depends on the data.


\subsection{Relations to Mahalanobis}
We discuss the generality of LSGM by comparing it to Mahalanobis.
Mahalanobis score is calculated as follow:
\begin{equation}
\label{eq:maha_ensemble}
S_{\mathrm{Maha}}(\bm{x})=\sum_i \alpha_{i} S^{i}_{\mathrm{Maha}}(\bm{x}),
\end{equation}
where parameter $\alpha_{i}$ is the weight of the $i$-th layer, which is learned by using a logistic regression on the \ood\ data.
The $S^i_{\mathrm{Maha}}(\bm{x})$ denotes the score of the $i$-th layer, which is calculated by
\begin{equation} \label{eq:maha}
S^{i}_{\mathrm{Maha}}(\bm{x})= \max_{j}- (\bm{x}_i-\bm{\mu}_i^j )^\mathsf{T}\bm{\Sigma}_i^{-1} (\bm{x}_i-\bm{\mu}_{i}^j ),
\end{equation}
where the $\bm{x}_i$ is the latent feature at the $i$-th layer, $\bm{\mu}_i^j$ and $\bm{\Sigma}_i$ are the $j$-th class mean and covariance matrix, respectively.
The score is calculated via Mahalanobis distance that how the data belongs to the nearest cluster in the selected representation space.
Mahalanobis selects a specific trace across representation spaces to execute a weighted summation upon the scores.
Consider the score in Eq.~\eqref{eq:maha},
assume the class $\bm{\mu}_i^j$ and covariance matrix $\bm{\Sigma}_i$ are mean and covariance matrix of a Gaussian distribution on $i$-th hidden layer,
the Mahalanobis score is the log-probability plus a fixed noise.
 Thus, Mahalanobis could be seen as a special case of LSGM in terms of the specific trace of occurring the nearest cluster on each representation space. By doing this, given an observed sequence of latent features $\bm{x}_{1}$, $\bm{x}_{2}$, $\dots$, $\bm{x}_{m}$, the hidden variables $\bm{z}$ are also specified, which makes the trace deterministic instead of probabilistic,
\begin{small}
\begin{eqnarray*}\label{eq:joint_decomposed}
\begin{aligned}
&\mathrm{LSGM}_{Maha}(\bm{x})\\
&\quad =\log\prod_{i=1}^{m} P(\bm{x}_{i}\mid z_{i}=nearest)^{\alpha_i}\\
&\quad =\sum_{i=1}^m\alpha_i( -\log\sqrt{(2\pi|\bm{\Sigma}_i|)} -
\frac{1}{2}\left(\bm{x}_{i}-\bm{\mu}_i^{z_i})^\mathsf{T}\bm{\Sigma}_i^{-1}(\bm{x}_{i}-\bm{\mu}_i^{z_i}) \right)\\
&\quad =-\frac{1}{2}\sum_{i=1}^m\alpha_i(\bm{x}_{i}-\bm{\mu}_i^{z_i})^\mathsf{T}\bm{\Sigma}_i^{-1}(\bm{x}_{i}-\bm{\mu}_i^{z_i})+\epsilon\\
&\quad =-\frac{1}{2}S_{\mathrm{Maha}}(\bm{x})+\epsilon \sim S_{Maha}(\bm{x})
\end{aligned}
\end{eqnarray*}
\end{small}
where $\alpha_i$ is the coefficient for each probability.

   \section{Experiments}

In this section, we demonstrate the effectiveness of the proposed method on several computer vision benchmark datasets. All experiments run on Linux with PyTorch. The code will be released for reproducibility.

\begin{table*}[htbp]
	\centering
	\renewcommand\arraystretch{0.8}
	\smallSize
	\caption{Detecting in-distribution and \ood\ test set data for the task of image classification.} 
	\label{table-ood-results}
	\begin{threeparttable}
	\begin{tabular}{p{1cm}<{\centering}p{2cm}<{\centering}p{2cm}<{\centering}p{3cm}<{\centering}p{3cm}<{\centering}p{3cm}<{\centering}}
		\toprule  
        \textbf{Model} & \textbf{In-distribution}&\textbf{OOD} &\textbf{TNR (at 95\% TPR)} &\textbf{AUROC} & \textbf{AUPR}\\
        \midrule
        & \multicolumn{4}{c}{\textbf{Methods: Softmax/ODIN*/Mahalanobis/Mahalanobis*/Generalized ODIN/LSGM}}\\
        \cmidrule{1-6}
        \multirow{14}{*}{ResNet} & \multirow{7}{*}{CIFAR-10} & Gaussian Noise & 76.9/49.7/90.4/99.7/95.8/\textbf{100} & 89.7/77.6/97.2/99.9/98.9/\textbf{100} & 82.4/65.6/95.8/99.8/98.4/\textbf{100}\\
        &  & Rademacher Noise & 85.6/88.6/97.7/100/99.9/\textbf{100} & 93.6/93.8/99.3/100/100\textbf{100} & 87.6/85.3/98.8/100/100\textbf{100}\\
        &  & Blob & 83.0/56.0/96.8/99.1/96.8/\textbf{99.8} & 93.7/80.3/99.4/99.7/99.0/\textbf{100} & 90.7/72.2/99.4/99.6/98.4/\textbf{99.9}\\
        &  & Texture & 65.4/42.6/79.9/56.8/\textbf{82.3}/76.4 & 89.2/79.8/95.2/96.1/91.4/\textbf{96.2} & 78.2/65.4/90.3/94.1/88.0/\textbf{95.0}\\
        &  & LSUN & 71.7/56.4/85.9/95.7/84.1/\textbf{97.0} & 91.1/81.0/96.6/99.0/96.5/\textbf{99.3} & 88.6/71.4/95.8/99.0/96.1/\textbf{99.3}\\ 
        &  & iSUN & 71.9/56.5/85.5/94.3/83.2/\textbf{95.9} & 91.0/81.7/96.3/98.8/96.4/\textbf{99.0} & 87.3/70.3/95.0/98.7/95.6/\textbf{99.0}\\
        &  & TinyImagenet & 71.6/56.6/80.7/\textbf{94.0}/70.4/92.7 & 91.0/82.4/95.0/\textbf{98.7}/94.2/98.6 & 88.3/73.3/94.0/98.7/94.1/\textbf{98.7}\\ 
        \cmidrule{2-6}
         & \multirow{7}{*}{CIFAR-100} & Gaussian Noise & 50.7/65.7/57.4/99.6/98.9/\textbf{100} & 66.8/85.3/65.1/99.8/99.4/\textbf{100} & 54.4/74.8/52.7/99.4/97.7/\textbf{100}\\
         &  & Rademacher Noise & 53.8/85.4/77.2/100/99.8/\textbf{100} & 69.8/94.3/88.1/100/99.9/\textbf{100} & 56.9/90.0/78.2/100/99.8/\textbf{100}\\
         &  & Blob & 70.1/63.3/78.7/94.7/95.5/\textbf{98.6} & 90.0/90.9/90.6/98.4/98.8/\textbf{99.5} & 86.5/89.5/84.6/97.4/98.4/\textbf{98.9}\\
         &  & Texture & 37.2/21.8/39.5/66.7/\textbf{70.8}/58.5 & 77.9/77.5/81.2/92.1/92.5/\textbf{92.9} & 62.5/67.8/68.4/87.5/86.4/\textbf{90.7}\\
         &  & LSUN & 35.3/35.9/42.0/81.3/77.2/\textbf{87.0} & 75.6/84.8/79.0/95.9/93.8/\textbf{97.0} & 71.7/85.0/74.1/95.6/91.7/\textbf{96.6}\\ 
         &  & iSUN & 36.7/36.4/41.6/77.5/75.3/\textbf{84.4} & 75.6/85.2/78.5/95.3/93.5/\textbf{96.5} & 68.7/83.9/71.0/94.9/90.6/\textbf{96.0}\\
         &  & TinyImagenet & 41.0/45.5/41.6/79.8/77.4/\textbf{82.1} & 77.1/87.5/78.9/96.1/94.2/\textbf{96.4} & 73.2/87.2/74.5/96.2/92.4/\textbf{96.5}\\ 
        \midrule
        \multirow{14}{*}{DenseNet} & \multirow{7}{*}{CIFAR-10} & Gaussian Noise & 93.7/96.8/100/100/99.9/\textbf{100} & 97.7/98.5/100/100/99.9/\textbf{100} & 96.4/95.3/100/100/99.5/\textbf{100}\\
        &  & Rademacher Noise & 93.1/98.0/100/100/99.9/\textbf{100} & 97.0/99.0/100/100/100/\textbf{100} & 94.1/97.1/100/100/99.9/\textbf{100}\\
        &  & Blob & 0.00/14.1/99.2/98.7/95.8/\textbf{99.6} & 64.0/51.2/99.7/99.2/98.2/\textbf{99.8} & 56.1/46.7/99.3/97.9/95.2/\textbf{99.4}\\
        &  & Texture & 60.2/8.82/67.6/56.4/\textbf{82.5}/80.8 & 88.5/68.3/95.2/92.2/96.2/\textbf{96.8} & 78.3/58.4/93.9/89.7/93.2/\textbf{95.6}\\
        &  & LSUN & 85.4/80.2/81.1/91.6/95.1/\textbf{95.7} & 95.5/93.5/96.4/98.0/98.6/\textbf{99.0} & 94.1/89.9/96.6/97.8/97.8/\textbf{99.0}\\ 
        &  & iSUN & 83.3/77.7/75.2/88.8/\textbf{94.9}/93.0 & 94.8/92.8/95.6/97.6/\textbf{98.7}/98.6 & 92.5/88.0/95.6/97.2/97.9/\textbf{98.6}\\
        &  & TinyImagenet & 81.2/71.1/72.1/89.3/\textbf{93.7}/91.2 & 94.1/91.3/95.4/97.7/\textbf{98.4}/98.3 & 92.4/87.6/95.9/97.6/97.8/\textbf{98.5}\\ 
        \cmidrule{2-6}
         & \multirow{7}{*}{CIFAR-100} & Gaussian Noise & 70.8/96.8/100/100/99.8/\textbf{100} & 86.6/98.4/100/100/99.9/\textbf{100} & 77.8/95.7/100/100/99.5/\textbf{100}\\
         &  & Rademacher Noise & 40.6/85.7/100/100/99.9/\textbf{100} & 52.0/91.7/100/100/99.9/\textbf{100} & 45.6/81.5/100/100/99.9/\textbf{100}\\
         &  & Blob & 64.0/82.5/96.4/97.2/96.6/\textbf{99.3} & 88.0/94.8/98.2/98.4/99.1/\textbf{99.7} & 84.7/93.1/94.8/95.3/98.4/\textbf{98.8}\\
         &  & Texture & 23.2/19.6/42.0/32.0/62.3/\textbf{63.8} & 72.6/73.0/90.0/85.5/90.9/\textbf{93.6} & 57.5/57.2/87.7/82.2/85.1/\textbf{91.4}\\
         &  & LSUN & 29.7/38.4/85.4/83.0/89.1/\textbf{90.6} & 70.8/79.5/96.7/96.1/97.4/\textbf{97.7} & 67.9/75.8/96.2/95.6/96.9/\textbf{97.5}\\ 
         &  & iSUN & 27.3/35.5/79.0/78.1/87.3/\textbf{88.0} & 69.6/78.5/95.9/95.4/97.3/\textbf{97.3} & 63.9/72.3/95.5/94.7/96.6/\textbf{97.0}\\
         &  & TinyImagenet & 27.9/36.7/77.0/79.3/\textbf{89.5}/84.7 & 71.6/80.3/95.8/95.8/\textbf{97.6}/97.1 & 69.0/77.3/95.9/95.7/97.2/\textbf{97.2}\\ 
         \midrule
         \multirow{7}{*}{WRN} & \multirow{7}{*}{TinyImageNet} & Gaussian Noise & 49.3/27.2/93.5/94.7/56.1/\textbf{98.4} & 63.7/48.0/95.6/96.3/86.6/\textbf{98.7} & 52.2/43.9/87.8/88.9/85.0/\textbf{95.0} \\
          &  & Rademacher Noise & 32.1/51.9/98.9/\textbf{99.6}/84.9/99.3 & 42.4/62.5/99.2/\textbf{99.7}/93.2/99.5 & 41.5/51.2/97.1/\textbf{98.7}/86.1/97.8\\
          &  & Blob & 38.0/59.8/97.8/98.5/96.0/\textbf{99.2} & 66.4/84.6/99.0/99.1/98.7/\textbf{99.4} & 56.1/76.8/97.1/97.0/98.0/\textbf{97.9}\\
          &  & Texture & 21.4/18.2/35.8/60.9/39.2/\textbf{61.8} & 66.2/72.7/85.2/91.4/83.5/\textbf{92.4} & 48.7/61.4/81.1/87.1/76.7/\textbf{89.1}\\
          &  & LSUN & 32.4/34.4/14.0/28.5/26.0/\textbf{44.2} & 73.7/75.0/57.0/64.8/67.8/\textbf{77.4} & 69.0/\textbf{69.5}/51.6/56.0/62.8/67.8\\
          &  & iSUN & 44.3/55.3/31.2/63.8/34.8/\textbf{81.5} & 80.1/86.1/73.8/87.6/74.4/\textbf{94.2} & 74.4/82.1/65.1/81.3/69.0/\textbf{91.3}\\
          &  & CIFAR-10 & 43.8/74.3/23.0/75.6/56.4/\textbf{90.8} & 81.9/94.2/65.3/89.4/89.3/\textbf{97.0} & 80.1/93.6/56.9/81.2/88.7/\textbf{95.0} \\

		\bottomrule
	\end{tabular}
\end{threeparttable}
\end{table*}

\subsection{Experimental Setup}
\textbf{Networks configurations:} In our experiments, we use three popular deep neural networks, ResNet-34, DenseNet-BC (\textit{L=100, k=12, droprate=0}), and WRN (\textit{L=28, widen-factor=2, droprate=0.3}) for evaluation.
ResNet-34 and WRN are the classic implementations from their original papers~\cite{he2016deep, Zagoruyko2016WRN}. The settings of DenseNet-BC are following the same setup from Mahalanobis\footnote{https://github.com/pokaxpoka/deep\_Mahalanobis\_detector}.

\textbf{In-distribution datasets:} The experiments involve three in-distribution datasets, CIFAR-10, CIFAR-100, and the large-scale dataset TinyImageNet. 
The models in experiments were also trained on these datasets.

\textbf{Out-of-distribution datasets:}  
We choose nine benchmark \ood\ datasets, including
three noises (\eg, Gaussian and Rademacher noises, and Blob) and six real-world datasets (\eg, Texture, LSUN, iSUN, and TinyImageNet). 


\textbf{Evaluation metrics:} We apply the three most widely used metrics in the previous work to measure the effectiveness of \ood\ detection.
The first one is the true negative rate (TNR) at 95\% true positive rate (TPR), which could be interpreted as the probability that an \ood\ input is correctly identified when the TPR is as high as 95\%.
The second one is the area under the receiver operating characteristic curve (AUROC)~\cite{davis2006relationship}.
The AUROC could be interpreted as the model's ability to discriminate between positive and negative inputs.
The third metric is the area under the precision-recall curve (AUPR).

\textbf{Compared methods:} We use five existing methods proposed for \ood\ detection, \ie, the baseline method~\cite{hendrycks2017baseline}, ODIN* (without \ood)~\cite{liang2018enhancing, hsu2020generalized}, Mahalanobis (penultimate layer)~\cite{lee2018simple} and its modified version Mahalanobis* (without \ood)~\cite{hsu2020generalized}, and Generalized ODIN (DeConf-C*)~\cite{hsu2020generalized}.
All methods are built without any \ood\ data.

\textbf{Hyperparameters:} Generative model does not need for each \ood\ dataset.
We focus on selected layers and the number of components \textit{k} for building LSGM. In this experiment, we use \textit{k=50} for models on CIFAR-10, \textit{k=100} for models on CIFAR-100, and \textit{k=200} for the model on TinyImageNet in terms of the quantities of classes in datasets. 
To show the simplicity, we select the last layer of each main block of the model's architecture for LSGM. No special selection strategy is applied in the evaluation.

%
%

\subsection{Evaluation Results}

\subsubsection{Out-of-distribution detection}
We evaluate the performance of LSGM compared to other methods built without \ood\ data. 
The training data is the training set used for building the DNN model. All test sets of these benchmark datasets are treated as in-distribution and \ood\ test sets, with 10,000 samples each.
To keep the functionality of input preprocessing used in Mahalanobis and ODIN, we adopt a modified input preprocessing strategy, which is introduced in Generalized ODIN.
Notice that No input preprocessing is applied for the proposed LSGM.
An overall comparison is detailed in Table~\ref{table-ood-results}. Mahalanobis and Generalized ODIN both outperform the softmax and ODIN*, and the proposed LSGM is better than them in most cases.
The results show strong evidence of the efficacy of \ood\ detection from generative manner.
In particular, our method is the dominant detector for detecting noises, winning 14 of 15 cases in terms of AUROC. Most of the results are 100\%, which means a zero error for noise detection.

For \ood\ real-world datasets, the LSGM outperforms the five compared methods, and is the most stable detector for different \ood\ data.
Note that, Generalized ODIN needs to retrain the DNN model.
In evaluation, the top-1 classification accuracies of retrained DNN models drop up to $3.74\%$ according to the difficulty of classification task, which makes it less practical for \ood\ detection on the DNN models in operation.

Overall, LSGM significantly outperforms these SOTA methods, wining 30 of 35 cases in terms of AUROC, which makes it be a stable method without any assistance of \ood\ data, showing the good generalization for the detection.

\subsubsection{Ablation study}

We study the influence of number of components and selected layers for LSGM to the detection performance.

\begin{figure}[t]
  \centering
  \subfigure[ResNet-34 on CIFAR-10]{
  \label{fig:k_cifar10_res}\centering
    \includegraphics[scale =0.43]{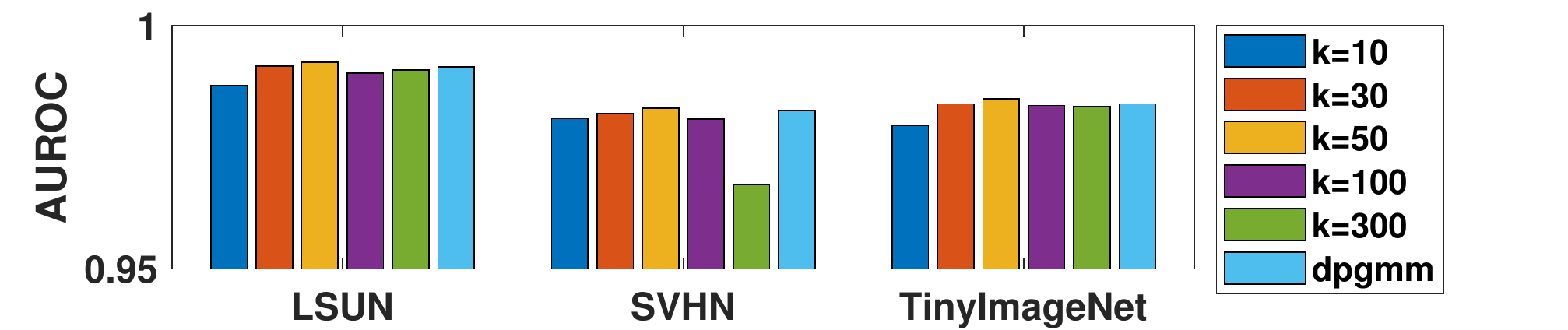}}
    \subfigure[DenseNet on CIFAR-10]{
      \label{fig:k_cifar10_dense}\centering
        \includegraphics[scale =0.43]{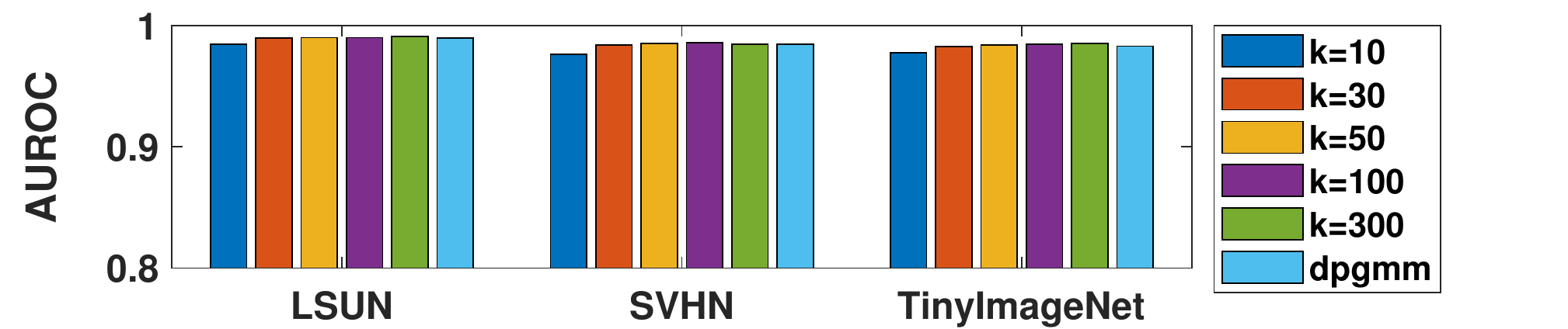}}
  \caption{Analysis on the influence on \#components for LSGM in representation spaces. } 
\label{fig:k_components}
\end{figure}

\textbf{Number of components:} 
Since we apply Gaussian Mixture Model in each representation space, the number of components for Gaussian Mixture Model is one of the key hyperparameters of the proposed LSGM.
To analyze the impact of \#components to \ood\ detection, we evaluate multiple LSGMs on ResNet-34 and DenseNet for both CIFAR-10 datasets with different \#components settings. 
Despite the manual selection of \#components, we also evaluate Dirichlet Process prior as an automated selection approach for \#components.
The results are shown in Fig.~\ref{fig:k_components}.
In general, the performance of \ood\ detection is steady with the increased number of components, with $\text{LSGM}_{k=300}$ meeting the highest overall performance in most cases.
However, $\text{LSGM}_{k=300}$ shows the worst AUROC on SVHN, indicating the lower separability with excessive amount of components for this specific \ood\ data. 
Dirichlet Process Gaussian Mixture model (DPGMM) shows the great performance in all cases, which proves the efficacy of searching the true active components in an adaptive manner.
As a result, we suggest DPGMM as the default approach to learn the distributions in representation spaces for LSGM.

\begin{figure}[t]
  \centering
    \subfigure[ResNet-34 on CIFAR 100]{
      \label{fig:m_cifar10_res}\centering
        \includegraphics[scale =0.23]{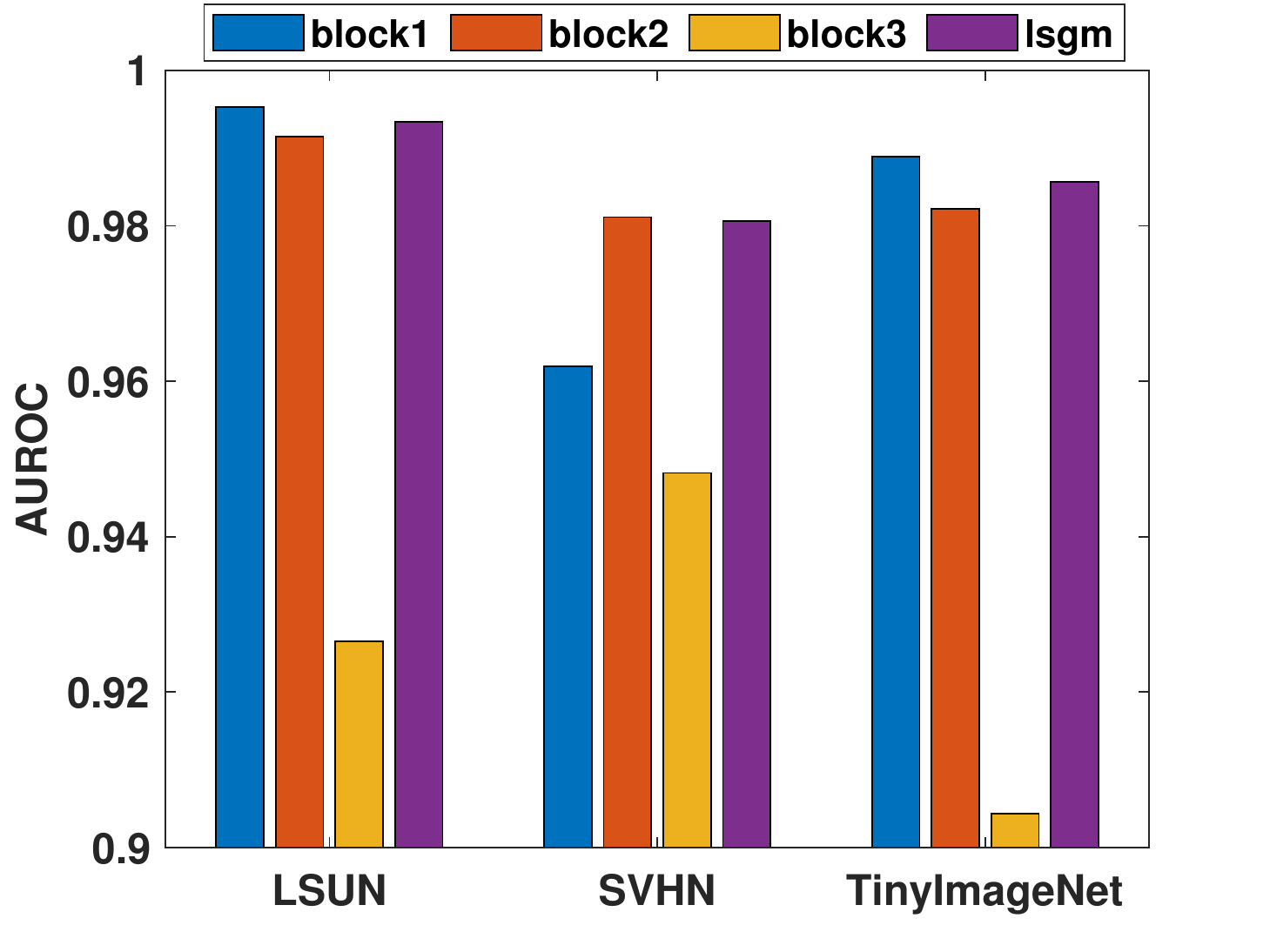}}
        \hspace{1em}
    \subfigure[DenseNet on CIFAR 100]{
      \label{fig:m_cifar10_dense}\centering
        \includegraphics[scale =0.23]{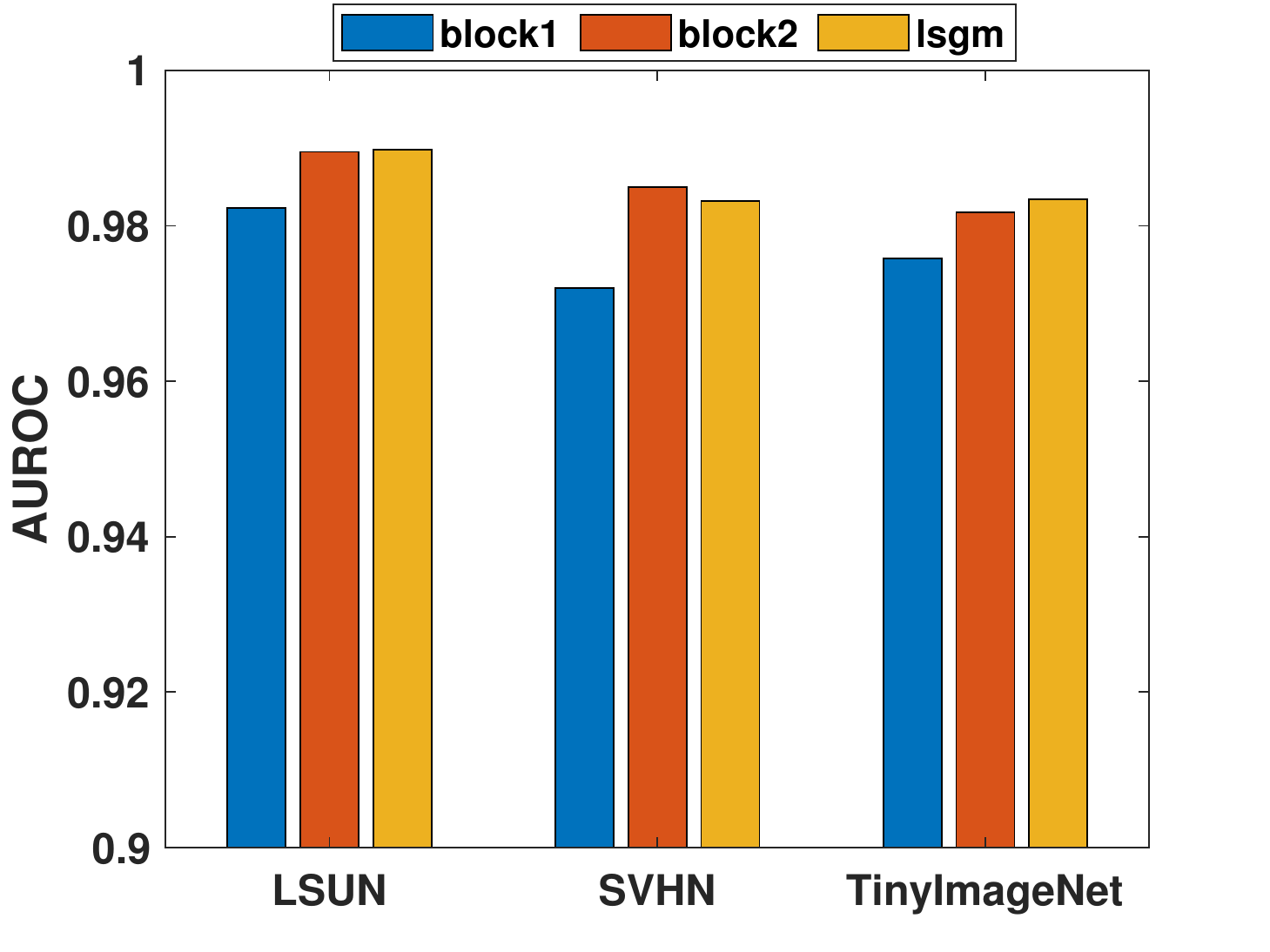}}
  \caption{Analysis on the influence on layer selection for LSGM. 
} 
\label{fig:m_layers}
\end{figure}

\textbf{Impact of layer selection:} LSGM consider the inference trace, so the strategy of layer selection is another important thing of the LSGM.
To analyze the impact, we also evaluate LSGM on ResNet-34 and DenseNet with three \ood\ datasets for clarity.
We compare the results from Gaussian Mixture Model in each hidden layer and the LSGM, which simply considers all selected layers in series.
As the results shown in Fig.~\ref{fig:m_layers}, it is hard to determine which layer has the best for all \ood\ data, since the performance of each single layer varies.
In contrast, LSGM shows the stable and well performance without the anxiety of layer selection.

\subsubsection{Time cost}

\begin{figure}[t]
 \centering
 \subfigure[Evaluation on different number of components.]{
 \label{fig:timecost_compoents}\centering
   \includegraphics[width=0.23\textwidth]{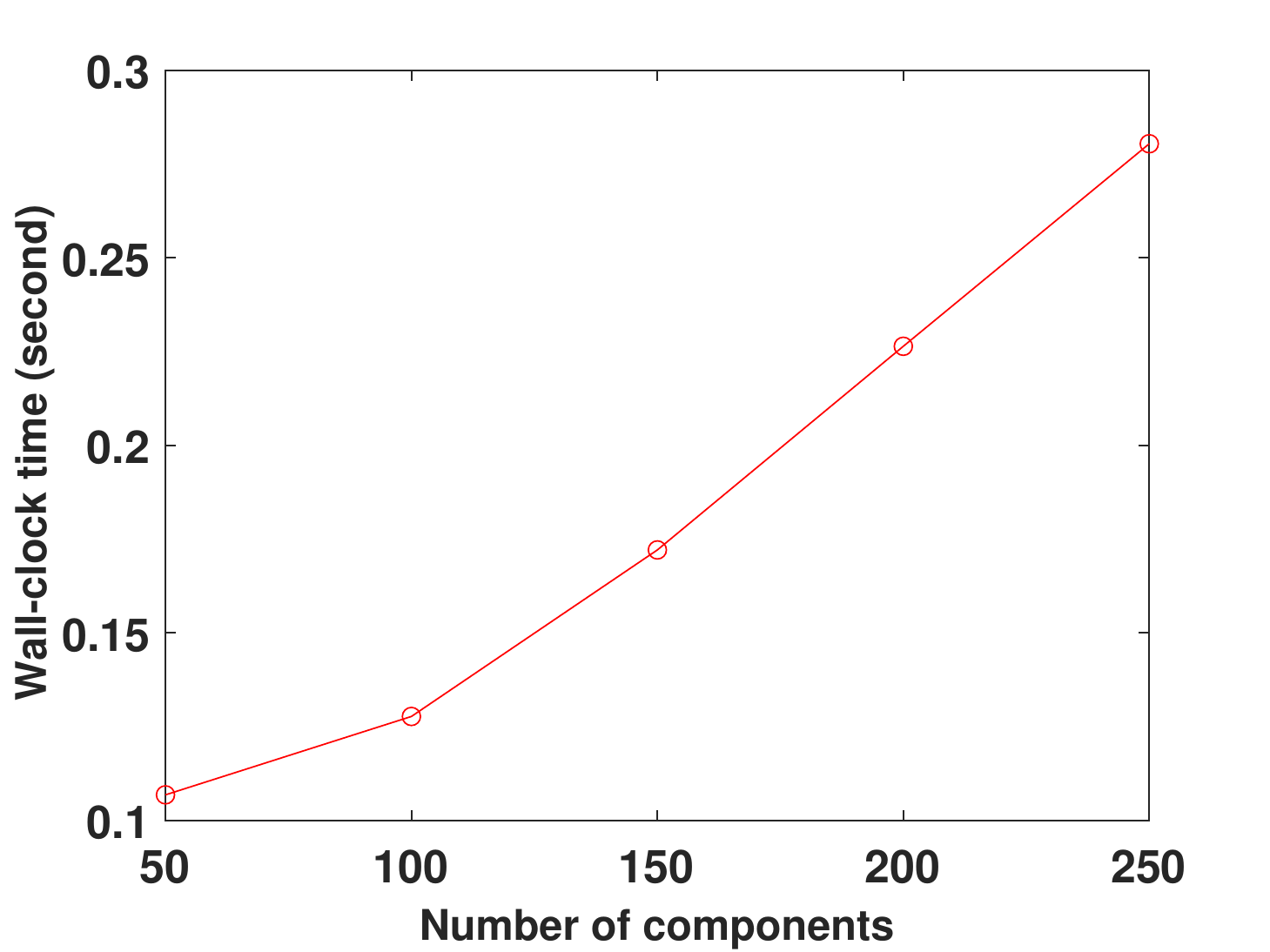}}
 \subfigure[Evaluation on different number of selected layers.]{
 \label{fig:timecost_layers}\centering
   \includegraphics[width=0.23\textwidth]{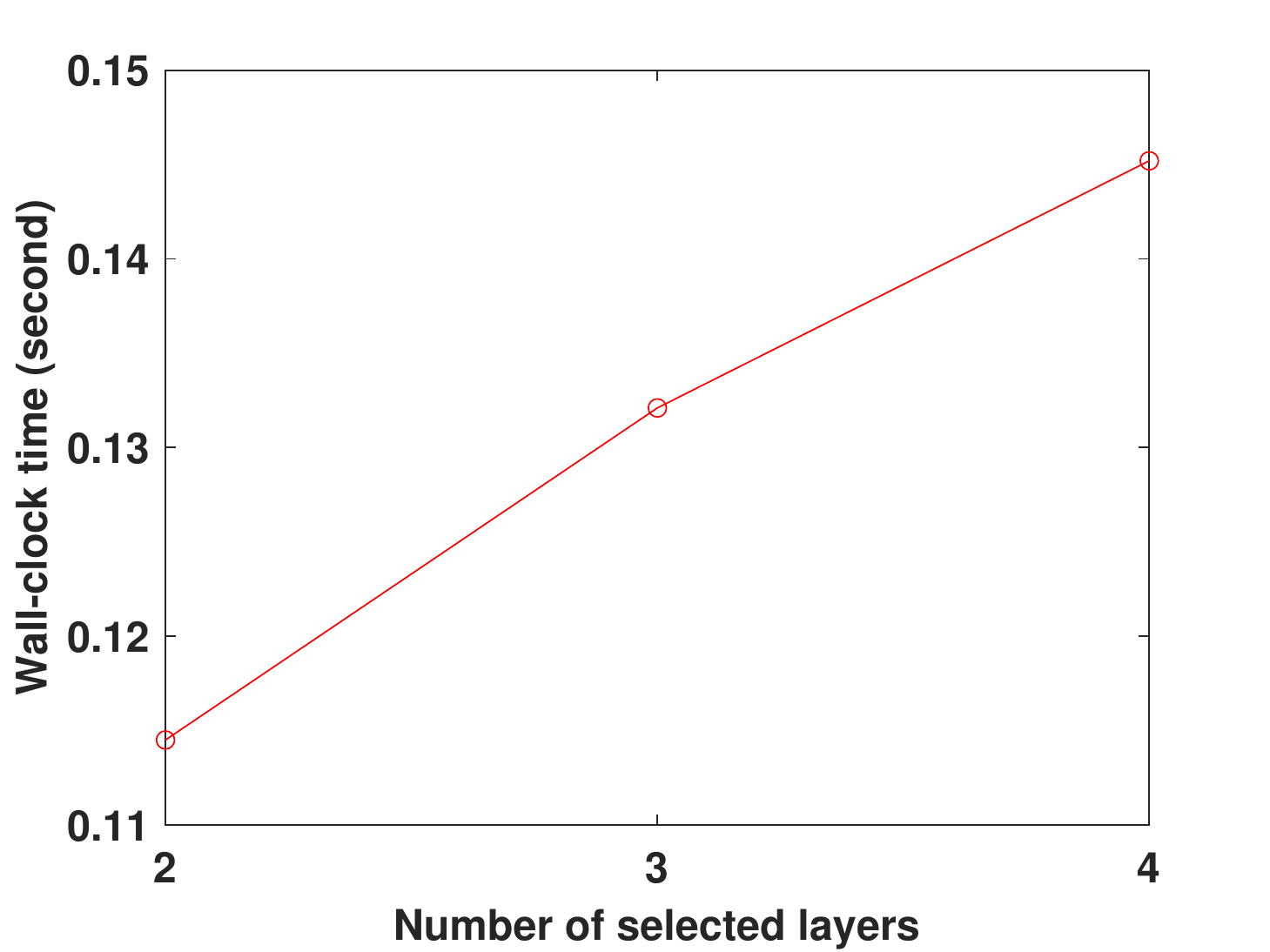}}
 \caption{Time cost evaluation for LSGM applied to WRN model trained on TinyImageNet. The results are average wall-clock time of \ood\ detection for each input. All numbers of components in hidden layers are the same.} 
\label{fig:timecost}
\end{figure}

In this part, we present the results of the proposed LSGM in terms of efficiency.
The experiments are run on a Linux server with two Intel Xeon Gold 5118 CPUs @2.30GHz, 10
GeForce RTX 2080Ti GPUs, and 384GB RAM, running Ubuntu 16.04.
In Fig.~\ref{fig:timecost}, we show the efficiency of LSGM on WRN model, which is trained on TinyImageNet dataset.
The WRN contains four main blocks in the architecture. We follow the strategy of hidden layer selection to uniformly select the last layer in each main block.
The number of selected layers is from $2$ to $4$.
We vary number of components $K_i$ in range $[50, 250]$ with the interval $50$, while fixing $m=3$.
The sizes of training and testing set are $100,000$ and $10,000$, respectively, and the batch size is $100$.
The results are the wall-clock time of the whole process of training and testing.
As shown in Fig.~\ref{fig:timecost_compoents} and Fig.~\ref{fig:timecost_layers}, the proposed LSGM scales linearly when number of components or selected layers increases, respectively.

   \section{Conclusions}
   In this paper, we propose a generative probabilistic graphical model across representation spaces, Latent Sequential Gaussian Mixture, to depict the process of DNN inference. The Gaussian Mixture model is used to present the distribution of latent features in each representation space.
As a new generative paradigm, LSGM is the first one to explicitly represent the inference via joint spaces in both philosophy and method perspectives.
Thus, \ood~detection could be solved steadily without the help of any \ood~data.
Our comprehensive evaluation shows that LSGM is effective and outperforms the compared methods.

{\small
\bibliographystyle{ieee_fullname}
\bibliography{iccv}
}

\end{document}